# Regularization of the movement of a material point along a flat trajectory: application to robotics problems

B.G. Mukanova,  M.A. Akhmetzhanov, D. N. Azimova

**Abstract.** The control problem of the working tool movement along a predefined trajectory is considered. The integral of kinetic energy and weighted inertia forces for the whole period of motion is considered as a cost functional. The trajectory is assumed to be planar and defined in advance. The problem is reduced to a system of ordinary differential equations of the fourth order. Numerical examples of solving the problem for movement along straight, circular and elliptical trajectories are given.

Keywords: numerical computing, optimal velocities, law of motion, predefined trajectory, minimal inertia, minimal kinetic energy, robots, 2dof.

Optimization problems of the movement of robots' working tools have many different formulations. Part of the formulations concerns determining the optimal trajectory, see, for instance the review [1] and references therein. Another directions of research concerns improvement of different performance indices [2, and references therein]. Here some key mathematical expressions of the indices are provided as well. Large number of articles are devoted to the kinematic performance evaluation and optimization as well, for instance, articles [4]-[11]. To improve the quality of work and control of the robot there are also problems of accurate positioning and following the trajectory [12, and references therein].

But compared to control of kinematic indices, the study of dynamic performances indices are developed slowly because of high diversity of robots and complexity of their dynamic models. However, because of wide applications of robots in recent years, the study of the dynamic performance indices is increasing [13]-[16].

We consider here a method of constructing the optimal law of motion along a given trajectory with minimizing kinetic energy and inertia forces for a given time interval in the case of 2dof. The relevance of this problem is due to the fact that in many technological problems the trajectory of the working tool is known in advance and depends on the operation performed by the robot.

Planar movements of the working tool are implemented by various robots, both sequential and parallel structure. An example is robot of the DexTAR type (dexterous twin-arm robot) [17], [18], as well as parallel robots with three degrees of freedom of 3RPR type. Several types of planar robots are described in Merlet monograph [19]. The 3D printers can be partially considered as planar robots as well, cause the printer head is moving mostly in a horizontal plane and is followed by the movement of the platform along vertical axis.

Let us consider a planar movement of some working tool of the robot. The following assumptions are accepted:

A) weight of supporting rods and other movable structural elements of the robot is small compared to the weight of the working tool (or platform with a working tool),

B) the weight of the working tool is concentrated near the center of gravity, or the working tool performs only translational motion,

C) the possible changes in the shape of the working tool in the movement process are neglected.

These hypotheses allow us to distract from the specific design of the robot and consider the movement of the working tool as the movement of a material point, or within the framework of the theory of translational motion of a solid body.

Let the Oxy plane of the Cartesian coordinate system coincides with the plane of the robot's movement. Let the position of the working tool is defined by the coordinates of the center of mass ($x_C$, $y_C$) in the motion plane.

Let the trajectory of the center of gravity of the moving platform be known and is given in the coordinate plane *Oxy* by the following parameterization

$$x_C = x(p),$$
$$y_C = y(p). \tag{1}$$

For instance, in a straight line movement, the parameter p can be set equal to one of the coordinates; in a circle movement, it can be set equal to the rotation angle of the platform carrying the tool. In general case, the parameter can be selected based on any convenient way of the trajectory definition.

The formulation of control problem for optimal movements of a working tool depends on practical applications and definition the quality criteria for such a movement. Quality criteria can be expressed in terms of power consumption, inertia forces, accelerations, vibration amplitudes, dimensions of the working area, etc. under some specified restrictions.

We consider the following optimization problem:

It is required to find the law of change of the parameter p=p(t) as a function of time, so that the movement of the platform, determined by the relations

$$x_C(t) = x(p(t)),$$
$$y_C(t) = y(p(t)), \tag{2}$$

has the minimum average kinetic energy for the considered time interval and at the same time provides the minimum of inertia forces.

Such a formulation makes sense if the aim is to reduce the power consumption of the manipulator and at the same time it is necessary to obtain the law of motion along the trajectory, ensuring the movement of the platform with minimum of inertia.

Let us assume that the positions and velocities of the platform at the initial and final time moments are given, and the time scale is selected so that the time changes in the interval [0,1]. We also assume that all values are reduced to a dimensionless form, after selecting appropriate length and velocity scales.

Suppose that at the initial and final moments the values of the function p (t) are given:

$$p(0) = p_0,$$
$$p(1) = p_1, \tag{3}$$

and at that moments the velocities are zero, i.e. these points are stopping points of the movement:

$$x(p_0) = x_0, \quad x(p_1) = x_1,$$
$$y(p_0) = y_0, \quad y(p_1) = y_1,$$
$$\dot{x}(p_0) = 0, \quad \dot{x}(p_1) = 0, \quad (4)$$
$$\dot{y}(p_0) = 0, \quad \dot{y}(p_1) = 0.$$

Points over functions, as usual, denote time derivatives. Prime symbols over the functions $x(p)$, $y(p)$ denote differentiation with respect to the parameter p.

Let $\vec{a}$ be the acceleration of the platform points. The total inertia force is obtained by integrating across the platform:

$$F_i = -\iint_\Omega \rho \vec{a} dS,$$

where $\rho$ is the surface density of the material of the movable platform or tool.

Given hypothesis B) that the mass of the tool is mainly concentrated near the center of gravity, we can approximately write that the inertia force is equal to

$$F_i = -\iint_\Omega \delta(\vec{r} - \vec{r}_C) \vec{a} dS = -m\vec{a}_C,$$

where m is the mass of the tool. The same formula holds in the case of translational movement of the tool.

As a measure of the magnitude of inertia forces for the whole movement time, we consider the integral of the squared inertia force, taken over the time interval [0,1]:

$$\|F_i\|^2 = \int_0^1 m^2 a_C^2(t) dt = m^2 \int_0^1 \left( \ddot{x}_C^2 + \ddot{y}_C^2 \right) dt.$$

Now let us write the kinetic energy of the platform movement, within the framework of assumptions A) - B) and integrate it over time. It consists of the kinetic energy of the mass center and the rotational energy around that center:

$$E_k = \frac{1}{2} \int_0^1 m\left( \dot{x}_C^2(p(t)) + \dot{y}_C^2(p(t)) \right) dt + \frac{1}{2} \int_0^1 I \dot{\varphi}^2(t) dt.$$

Here $I$ is the moment of inertia, $\varphi(t)$ is the angular velocity of the tool.

Further for the sake of brevity, we omit the "C" index near the coordinates of the mass center.

According to our assumption, the whole mass is concentrated in the center of mass of the platform, therefore, the kinetic energy of the rotational motion can be neglected compared with the energy of the translational motion, then

$$E_k = \frac{1}{2} \int_0^1 m\left( \dot{x}^2(p(t)) + \dot{y}^2(p(t)) \right) dt.$$

The quality functional in the form of the sum of kinetic energy and weighed measure of inertia forces is defined as follows:

$$J(p) = \frac{1}{2}\int_0^1 m\left(\dot{x}^2(t) + \dot{y}^2(t)\right)dt + \frac{\alpha m^2}{2}\int_0^1 \left(\ddot{x}^2(t) + \ddot{y}^2(t)\right)dt. \tag{5}$$

Here α > 0 is a positive parameter that determines the weight of inertia forces in the cost functional (5). The larger the parameter α is, the greater be the influence of inertia forces on the cost functional.

***Formulation of the movement optimization problem***: *it is necessary to minimize the functional (5) with respect to the function p(t) that define the law of motion along the trajectory (x(t), y(t)) according to the formulas (1)-(2).*

Formula (5) is the quadratic norm of the two-dimensional vector function (x(t), y(t)) in the space $W_2^2[0,1]$ formed by functions with quadratically summed derivatives up to the second order.

It can be shown that this space is a Hilbert's space [20], [21] and functional (5), as the norm of a function in a Hilbert's space, is strongly convex [22]. However, we will consider it as a functional defined on a subset of quadratically summed functions $L_2[0,1]$. Then any bounded closed set from $W_2^2[0,1]$ is compact by virtue of the embedding theorems [5]. According to [22], the functional reaches its minimum on a convex compact set. Besides, if the functional is differentiable, then at the minimum point its Frechet derivative is equal to zero. Moreover, this condition is sufficient for the minimality of the cost functional.

To determine the Frechet derivative of the cost functional, let us first derive the formula for the first variation of the functional (5). We assume that all arguments p(t) and increments of the function δp(t) lie in some bounded convex set from $W_2^2[0,1]$. Then the increment of the functional (5) corresponding to the increment of the argument p is written as follows:

$$\delta J(p) = J(p + \delta p) - J(p) = \frac{m}{2}\int_0^1 \left(\dot{x}^2(p+\delta p) - \dot{x}^2(p) + \dot{y}^2(p+\delta p) - \dot{y}^2(p)\right)dt$$
$$+ \frac{\alpha m^2}{2}\int_0^1 \left(\ddot{x}^2(p(t)+\delta p(t)) - \ddot{x}^2(p(t)) + \ddot{y}^2(p(t)+\delta p(t)) - \ddot{y}^2(p(t))\right)dt. \tag{6}$$

Taking into account that

$$\dot{x} = \frac{dx(p)}{dp}\dot{p} = x'\dot{p}, \quad \dot{y} = \frac{dy(p)}{dp}\dot{p} = y'\dot{p}, \tag{7}$$

with first order of accuracy we have:

$$\delta \dot{x}^2(p) = 2\dot{x}\delta\dot{x}(p) = 2\dot{x}\left(x'(p+\delta p)(\dot{p}+\delta\dot{p}) - x'(p)\dot{p}\right) =$$
$$= 2\dot{x}\left((x'(p) + x''(p)\delta p)\dot{p} + x'(p+\delta p)\delta\dot{p} - x'(p)\dot{p}\right) = \tag{8}$$
$$= 2\dot{x}\left(x''(p)\dot{p}\delta p + x'(p)\delta\dot{p}\right),$$

i.e.

$$\delta\dot{x}^2(p) = 2\dot{x}\left(x''\dot{p}\delta p + x'\delta\dot{p}\right)$$

Similar formula is valid for increment $\delta\dot{y}^2(p)$. For the second derivative, we have

$$\ddot{x} = \frac{d}{dt}(x'\dot{p}) = x''(p)\dot{p}^2 + x'(p)\ddot{p},$$
$$\ddot{y} = y''(p)\dot{p}^2 + y'(p)\ddot{p}. \qquad (9)$$

Taking into account expressions (9), for the increment of the second derivative $\delta\ddot{x}^2(p)$, we obtain with first order of accuracy:

$$\delta\ddot{x}^2(p) = 2\ddot{x}\delta\ddot{x}(p) = 2\ddot{x}\left(\frac{d^2x(p+\delta p)}{dp^2}(\dot{p}^2 + \delta\dot{p}^2) + \frac{dx(p+\delta p)}{dp}(\ddot{p} + \delta\ddot{p})\right.$$
$$\left. -\frac{d^2x(p)}{dp^2}\dot{p}^2 - \frac{dx(p)}{dp}\ddot{p}\right) = 2\ddot{x}\left(x'''(p)\dot{p}^2\delta p + x''(p)(2\dot{p}\delta\dot{p} + \ddot{p}\delta p) + x'\delta\ddot{p}\right). \qquad (10)$$

The formula for increment of the second derivative $\delta\ddot{y}^2(p)$ is written similarly.

Taking into account formulas (8) - (10) for the linear part of the functional increment, we have:

$$\delta J(p) = m\int_0^1 \dot{x}(x''\dot{p}\delta p + x'\delta\dot{p}) + \dot{y}(y''\dot{p}\delta p + y'\delta\dot{p})dt$$
$$+\alpha m^2 \int_0^1 \ddot{x}\left(x'''\dot{p}^2\delta p + x''(2\dot{p}\delta\dot{p} + \ddot{p}\delta p) + x'\delta\ddot{p}\right)dt \qquad (11)$$
$$+\alpha m^2 \int_0^1 \ddot{y}\left(y'''\dot{p}^2\delta p + y''(2\dot{p}\delta\dot{p} + \ddot{p}\delta p) + y'\delta\ddot{p}\right)dt.$$

Considering functional (11) as a functional defined on some subset of quadratically summable functions $W_2^2[0,1]$ in $L_2[0,1]$, we represent its increment in the form

$$\delta J(p) = \langle J'(p), \delta p \rangle = \int_0^1 J'(p(t))\delta p(t)dt. \qquad (12)$$

Let the functions $x(p)$ and $p(t)$ be sufficiently smooth, so that all our derivatives are legitimate. It will be clear from the calculations for which smoothness representation (12) is acceptable. In this way, we determine on which subspace of functions from $L_2[0,1]$, we differentiate the functional (5) in the Frechet sense; then the condition of equality to zero of the Frechet derivative will be useful to find the minimum point.

Collecting the coefficients at $\delta p, \delta\dot{p}, \delta\ddot{p}$ in (11), we rewrite expression (11) in the following form:

$$\delta J(p) = \int_0^1 f_1(t)\delta p\,dt + \int_0^1 f_2(t)\delta\dot{p}\,dt + \int_0^1 f_3(t)\delta\ddot{p}\,dt \equiv J_1(p) + J_2(p) + J_3(p), \qquad (13)$$

where the functions $f_i(p), i = \overline{1,3}$ are defined by the formulas:

$$f_1(t) = m(\dot{x}x'' + \dot{y}y'')\dot{p} + \alpha m^2(\ddot{x}x''' + \ddot{y}y''')\dot{p}^2 + \alpha m^2(\ddot{x}x'' + \ddot{y}y'')\ddot{p},$$
$$f_2(t) = m(\dot{x}x' + \dot{y}y') + 2\alpha m^2(\ddot{x}x'' + \ddot{y}x'')\dot{p}, \quad (14)$$
$$f_3(t) = \alpha m^2(\ddot{x}x' + \ddot{y}x').$$

In formulas (14), we should substitute the time derivatives of the functions $x(t)$, $y(t)$ from formulas (7) and (9), which we did not do in order not to clutter the description.

Integrating by parts in (13), we obtain:

$$J_2(p) = \int_0^1 f_2(p)\delta\dot{p}\,dt = f_2(p)\delta p\Big|_{t=0}^{t=1} - \int_0^1 \dot{f}_2\delta p\,dt = -\int_0^1 \dot{f}_2\delta p\,dt,$$

$$J_3(p) = \int_0^1 f_3(p)\delta\ddot{p}\,dt = f_3(p)\delta\dot{p}\Big|_{t=0}^{t=1} - \int_0^1 \dot{f}_3\delta\dot{p}\,dt = -\dot{f}_3\delta p\Big|_{t=0}^{t=1} +$$
$$+ \int_0^1 \ddot{f}_3\delta p\,dt = \int_0^1 \ddot{f}_3(t)\delta p\,dt.$$

Hence

$$\delta J(p) = \left\langle \left(f_1(t) - \dot{f}_2(t) + \ddot{f}_3(t)\right), \delta p \right\rangle.$$

Then, if the functions $f_2(t), f_3(t)$ are sufficiently smooth, the functional J(p) is differentiable in the Frechet sense at the point p, and its gradient is given by the function

$$J'(p) = f_1(t) - \dot{f}_2(t) + \ddot{f}_3(t).$$

The equation that allows to determine the minimum point of the functional has the form

$$f_1(t) - \dot{f}_2(t) + \ddot{f}_3(t) = 0 \quad (15)$$

for boundary conditions following from formulas (3) and (4):

$$p(0) = p_0,\ \dot{p}(0) = 0,\ p(1) = p_1,\ \dot{p}(1) = 0.$$

For the convenience of applying mathematical packages, we reduce equation (15) to a system of quasilinear equations. For this aim, the following notations are used:

$$z_0(t) = p(t),$$
$$z_1(t) = \dot{p}(t),$$
$$z_2(t) = \ddot{p}(t), \quad (16)$$
$$z_3(t) = \dddot{p}(t).$$

Then equalities (7) and (9) can be rewritten in the form:

$$\dot{x} = x'(z_0(t))z_1(t),\ \dot{y} = y'(z_0(t))z_1(t),$$
$$\ddot{x} = x''(z_0(t))z_1^2(t) + x'(z_0(t))z_2(t), \quad (17)$$
$$\ddot{y} = y''(z_0(t))z_1^2(t) + y'(z_0(t))z_2(t).$$

In the future, we will omit the argument $(z_0(t))$ to shorten the representation. Next we need expressions for the following time derivative of third order:

$$\dddot{x} = x'''z_1^3 + 3x''z_1 z_2 + x'z_3,$$
$$\dddot{y} = y'''z_1^3 + 3y''z_1 z_2 + y'z_3,$$
(18)

Taking into account the notations (16) and expressions (17), the summand $f_1(t)$ in equation (15) has the form:

$$f_1(t) = m(\dot{x}x'' + \dot{y}y'')z_1(t) + \alpha m^2(\dddot{x}x''' + \dddot{y}y''')z_1^2(t) + \alpha m^2(\ddot{x}x'' + \ddot{y}y'')z_2(t) =$$
$$= (x'x'' + y'y'')(mz_1^2(t) + \alpha m^2 z_2^2(t)) + \alpha m^2 (x''x''' + y''y''')z_1^4(t) +$$
$$+ \alpha m^2 ((x'x''' + y'y''') + (x''^2 + y''^2))z_1^2(t)z_2(t).$$
(19)

Now calculate the time derivative of the function $f_2(t)$:

$$\dot{f}_2(t) = m\frac{d}{dt}(\dot{x}x' + \dot{y}y') + 2\alpha m^2 \frac{d}{dt}((\ddot{x}x'' + \ddot{y}y'')\dot{p}) =$$
$$= m(\ddot{x}x' + \ddot{y}y') + m(\dot{x}x'' + \dot{y}y'')\dot{p} + 2\alpha m^2(\ddot{x}x'' + \ddot{y}y'')\dot{p} +$$
$$+ 2\alpha m^2(\dddot{x}x''' + \dddot{y}y''')\dot{p}^2 + 2\alpha m^2(\ddot{x}x'' + \ddot{y}y'')\ddot{p}.$$
(20)

Let's write separately the components of expression (20) in terms of notations (16):

$$m(\ddot{x}x' + \ddot{y}y') + m(\dot{x}x'' + \dot{y}y'')\dot{p} = m((x''z_1^2 + x'z_2)x' + (y''z_1^2 + y'z_2)y') +$$
$$m(x'x''z_1 + y'y''z_1)z_1 = m(2(x''x' + y''y')z_1^2 + (x'^2 + y'^2)z_2),$$

$$(\ddot{x}x'' + \ddot{y}y'')\dot{p} + (\dddot{x}x''' + \dddot{y}y''')\dot{p}^2 = (x''x''' + y''y''')z_1^4 + 3(x''^2 + y''^2)z_1^2 z_2 +$$
$$(x'x'' + y'y'')z_1 z_3 + (x''x''' + y''y''')z_1^4 + (x'x''' + y'y''')z_1^2 z_2 =$$
$$= 2(x''x''' + y''y''')z_1^4 + (3(x''^2 + y''^2) + x'x''' + y'y''')z_1^2 z_2 + (x'x'' + y'y'')z_1 z_3,$$
(21)

$$(\ddot{x}x'' + \ddot{y}y'')\ddot{p} = (x''^2 z_1^2 z_2 + x'x''z_2^2) + (y''^2 z_1^2 z_2 + y'y''z_2^2) =$$
$$(x''^2 + y''^2)z_1^2 z_2 + (x'x'' + y'y'')z_2^2.$$

Now, taking into account expressions (21), we can rewrite the formula for the coefficient $\dot{f}_2(t)$ in equation (15)

$$\dot{f}_2(t) = m(2(x''x' + y''y')z_1^2 + (x'^2 + y'^2)z_2) +$$
$$2\alpha m^2 \Big(2(x''x''' + y''y''')z_1^4 + (4(x''^2 + y''^2) + x'x''' + y'y''')z_1^2 z_2 +$$
$$+ (x'x'' + y'y'')z_1 z_3 + (x'x'' + y'y'')z_2^2\Big).$$
(22)

We calculate the derivatives of the function $f_3(t)$ in (14):

$$\dddot{f}_3(t) = \alpha m^2 \frac{d^2}{dt^2}(\ddot{x}x' + \ddot{y}y'),$$

$$\frac{d}{dt}(\ddot{x}x' + \ddot{y}y') = \dddot{x}x' + \dddot{y}y' + \ddot{x}x''\dot{p} + \ddot{y}y''\dot{p},$$

$$\frac{d^2}{dt^2}(\ddot{x}x' + \ddot{y}y') = \ddddot{x}x' + \ddddot{y}y' + 2\dddot{x}x''\dot{p} + 2\dddot{y}y''\dot{p}$$
$$+ \ddot{x}x'''\dot{p}^2 + \ddot{y}y'''\dot{p}^2 + \ddot{x}x''\ddot{p} + \ddot{y}y''\ddot{p}.$$

Using the designations (16) and equalities (17), (18), we have:

$$\frac{d^2}{dt^2}(\ddot{x}x' + \ddot{y}y') = \ddddot{x}x' + \ddddot{y}y' + 2(\dddot{x}x'' + \dddot{y}y'')z_1 + (\ddot{x}x''' + \ddot{y}y''')z_1^2 + (\ddot{x}x'' + \ddot{y}y'')z_2 =$$
$$= \ddddot{x}x' + \ddddot{y}y' + 3(x''x''' + y''y''')z_1^4 + 7(x''^2 + y''^2)z_1^2 z_2 + 2(x'x'' + y'y'')z_1 z_3 + \quad (23)$$
$$+ (x'x''' + y'y''')z_1^2 z_2 + (x'x'' + y'y'')z_2^2.$$

In formula (23), we need formulas for the fourth time derivative of the coordinates $x(t)$ and $y(t)$. Differentiating (18), we get:

$$\ddddot{x} = \frac{d}{dt}\left(x'''z_1^3 + 3x''z_1 z_2 + x'z_3\right) = x^{(IV)}z_1^4 + 3x'''z_1^2 z_2 + x''z_1 z_3 +$$
$$3x'''z_1^2 z_2 + 3x''z_2^2 + 3x''z_1 z_3 + x'\dot{z}_3 = x^{(IV)}z_1^4 + 6x'''z_1^2 z_2 + 4x''z_1 z_3 + 3x''z_2^2 + x'\dot{z}_3. \quad (24)$$

The formulas for the fourth time derivative of the function $y(t)$ have a similar form.

Substitute (24) in (23):

$$\frac{d^2}{dt^2}(\ddot{x}x' + \ddot{y}y') = \left(x^{(IV)}z_1^4 + 6x'''z_1^2 z_2 + 4x''z_1 z_3 + 3x''z_2^2 + x'\dot{z}_3\right)x' +$$
$$\left(y^{(IV)}z_1^4 + 6y'''z_1^2 z_2 + 4y''z_1 z_3 + 3y''z_2^2 + y'\dot{z}_3\right)y' + \quad (25)$$
$$3(x''x''' + y''y''')z_1^4 + 7(x''^2 + y''^2)z_1^2 z_2 + 2(x'x'' + y'y'')z_1 z_3 +$$
$$+ (x'x''' + y'y''')z_1^2 z_2 + (x'x'' + y'y'')z_2^2.$$

Reducing similar terms in (25), we get:

$$\frac{d^2}{dt^2}(\ddot{x}x' + \ddot{y}y') = \left(x^{(IV)}x' + y^{(IV)}y' + 3(x''x''' + y''y''')\right)z_1^4 +$$
$$\left(x'^2 + y'^2\right)\dot{z}_3 + 7(x''^2 + y''^2 + x'x''' + y'y''')z_1^2 z_2 + 6(x'x'' + y'y'')z_1 z_3 + \quad (26)$$
$$+ 4(x'x'' + y'y'')z_2^2.$$

By substituting (22), (26), (19) into (15), we write equation (15) in expanded form in the notations (2) and (16):

$$(x'x''+y'y'')(mz_1^2+\alpha m^2 z_2^2)+\alpha m^2(x''x'''+y''y''')z_1^4+$$
$$+\alpha m^2\big((x'x'''+y'y''')+(x''^2+y''^2)\big)z_1^2 z_2$$
$$-m\big(2(x''x'+y''y')z_1^2+(x'^2+y'^2)z_2\big)$$
$$-2\alpha m^2\Big(2(x''x'''+y''y''')z_1^4+\big(4(x''^2+y''^2)+x'x'''+y'y'''\big)z_1^2 z_2$$
$$+(x'x''+y'y'')z_1 z_3+(x'x''+y'y'')z_2^2\Big)$$
$$+\alpha m^2(x^{(IV)}x'+y^{(IV)}y'+3(x''x'''+y''y'''))z_1^4$$
$$+\alpha m^2\big(x'^2+y'^2\big)\dot z_3+7\alpha m^2(x''^2+y''^2+x'x'''+y'y''')z_1^2 z_2+6\alpha m^2(x'x''+y'y'')z_1 z_3$$
$$+4\alpha m^2(x'x''+y'y'')z_2^2=0. \qquad (27)$$

By regrouping and collecting similar items in (27), we get:

$$\alpha m^2\big(x'^2+y'^2\big)\dot z_3-m\big(x'^2+y'^2\big)z_2+m(x'x''+y'y'')\big(-z_1^2+3\alpha m z_2^2\big)$$
$$+4\alpha m^2(x'x''+y'y'')z_1 z_3+6\alpha m^2(x'x'''+y'y''')z_1^2 z_2 \qquad (28)$$
$$+\alpha m^2\Big[(x^{(IV)}x'+y^{(IV)}y')+4(x''x'''+y''y''')\Big]z_1^4=0.$$

Equation (28) can be written in the form resolved with respect to $\dot z_3$:

$$\dot z_3=\frac{1}{\alpha m}z_2+\frac{(x'x''+y'y'')}{\alpha m(x'^2+y'^2)}z_1^2-\frac{(x'x''+y'y'')}{(x'^2+y'^2)}\big(4z_1 z_3+3z_2^2\big)$$
$$-\frac{6(x'x'''+y'y''')}{(x'^2+y'^2)}z_1^2 z_2-\frac{(x^{(IV)}x'+y^{(IV)}y')+4(x''x'''+y''y''')}{(x'^2+y'^2)}z_1^4. \qquad (29)$$

Definitions (16) can be written in the form of equations:

$$\dot z_0(t)=z_1(t),$$
$$\dot z_1(t)=z_2(t), \qquad (30)$$
$$\dot z_2(t)=z_3(t).$$

Equation (29) together with (30) forms a quasilinear system of ordinary differential equations, which has a canonical form.

The specific type of system (29) - (30) depends on the parameterization of the trajectory. Let's consider a few special cases. In the presented examples, the mass of the tool is taken as a unit of mass, the motion time is a time unit, and all values of length are also selected in a certain unit of length.

Example 1. Let the tool trajectory be straight and parameterized by the relations

$$x=p$$
$$y=kp+b=kx+b.$$

The case of a straight trajectory is important for practical applications, since it is often implemented in practice. For example, in 3D printing with filling of some plane area, the printer head moves along a set of parallel straight lines.

Obviously $x' = 1, y' = k$, the remaining derivatives are equal to zero. Therefore, equation (29) takes a simple form

$$\dot{z}_3 = \frac{1}{\alpha m} z_2$$

Returning to definitions (16), we have:

$$\dddot{p} = \frac{1}{\alpha m} \dot{p}$$

The solution to this problem is obtained in analytical form and is given by the formulas:

$$x(t) = p(t) = A\sinh(\gamma t) + B\cosh(\gamma t) + Ct + D,$$

where

$$\gamma = (\alpha m)^{-1/2}, A = \frac{(x_1 - x_0)\sinh(\gamma)}{\Delta}, B = \frac{(x_1 - x_0)(1 - \cosh(\gamma))}{\Delta},$$

$$\Delta = \sinh(\gamma)(\sinh(\gamma) - \gamma) - (\cosh(\gamma) - 1)^2 = 2(\cosh(\gamma) - 1) - \gamma\sinh(\gamma),$$

$$D = x_0 - B, C = -A\gamma$$

Figure 1 shows the optimal law of change of the abscissa for different parameters α. A trajectory of the form y=-2x+1 passing between points (0,1) and (1, -1) is used for calculations.

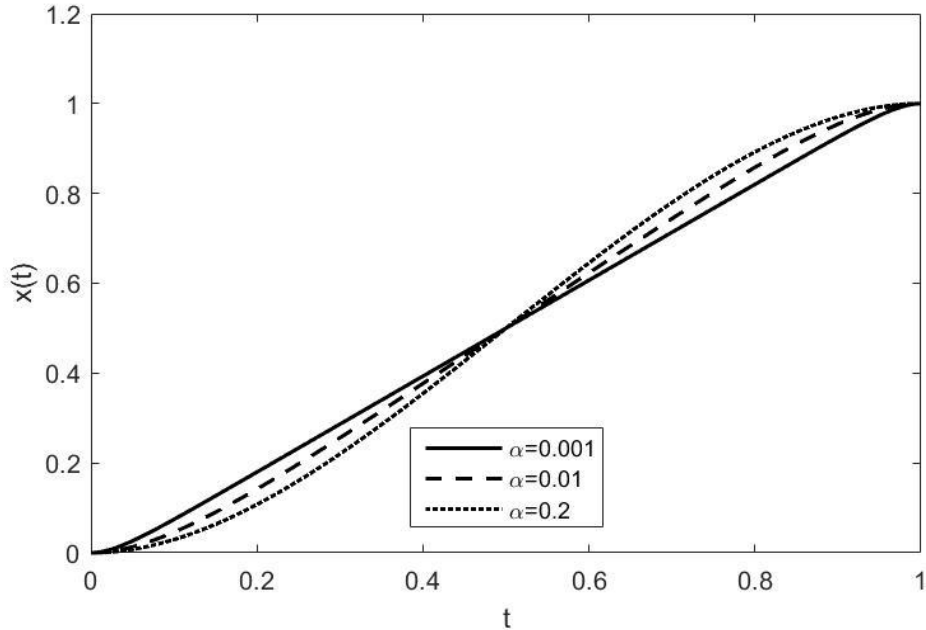

Figure 1 – the law of change of the abscissa at different weight coefficients α in the cost functional

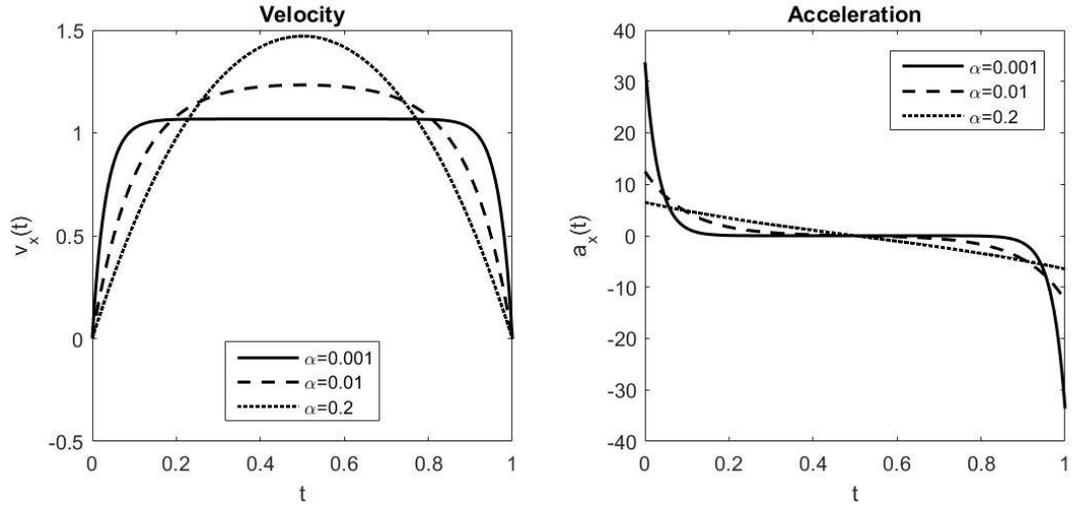

Figure 2- velocity and acceleration components on the Ox axis for straight-line movement along the trajectory at optimal motion mode

Figure 1 shows the calculation results — the law of the change in abscissa x(t) at different weight coefficients α in the quality functional. Figure 2 shows the velocities on the Ox axis at different weights α (left) and the acceleration on the Ox axis depending on the time for different α (right). It can be seen that the velocities change more smoothly at greater values of α.

Example 2. Circular motion. In this case the trajectory parameterization and parameter derivatives are set by the relations:

$$x = R\cos p, y = R\sin p,$$
$$x' = -R\sin p, y' = R\cos p,$$
$$x'' = -R\cos p, y'' = -R\sin p,$$
$$x''' = R\sin p, y''' = -R\cos p,$$
$$x^{(IV)} = R\cos p, y^{(IV)} = R\sin p.$$

Substitute these expressions in equation (29) and get:
$$\dot{z}_3 = \frac{1}{\alpha m} z_2 + 6 z_1^2 z_2.$$

In terms of definitions (16), we obtain a fourth-order quasilinear equation:

$$\dddot{p} = \left(\frac{1}{\alpha m} + 6\dot{p}^2\right)\ddot{p}.$$

The solution to this equation is obtained numerically. As an example, we considered the case of movement in a semicircle. In this case, the parameter describing the trajectory is the polar angle that changes in the interval $p_0 = 0, p_1 = \pi$.

The boundary conditions are set according to (4). The movement is made from a point with coordinates (R, 0) along a semicircle of radius R, to a point with coordinates (-R, 0). In the calculation, the length scale was taken equal to R.

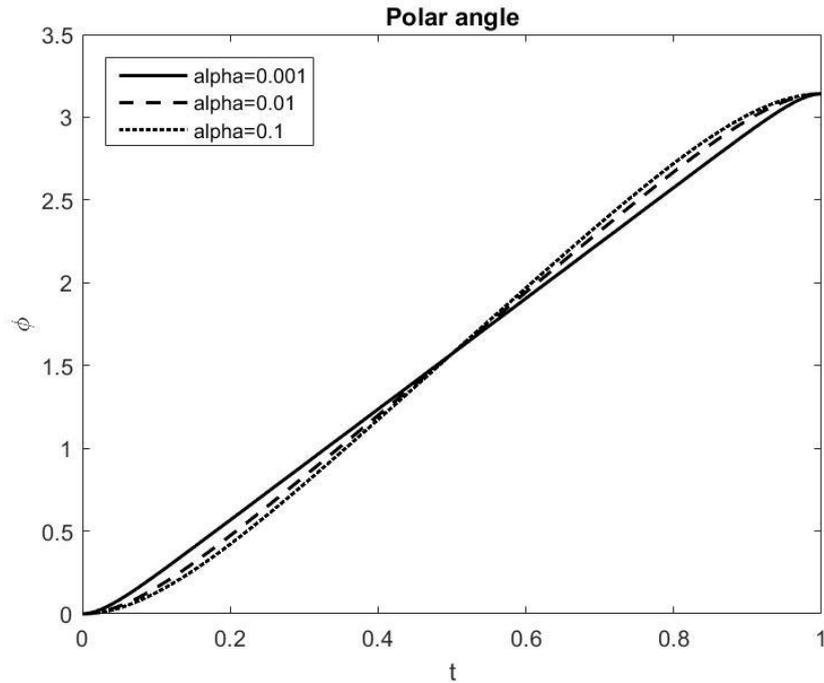

Figure 3 – The optimal function of changing the rotation angle for movement in a semicircle.

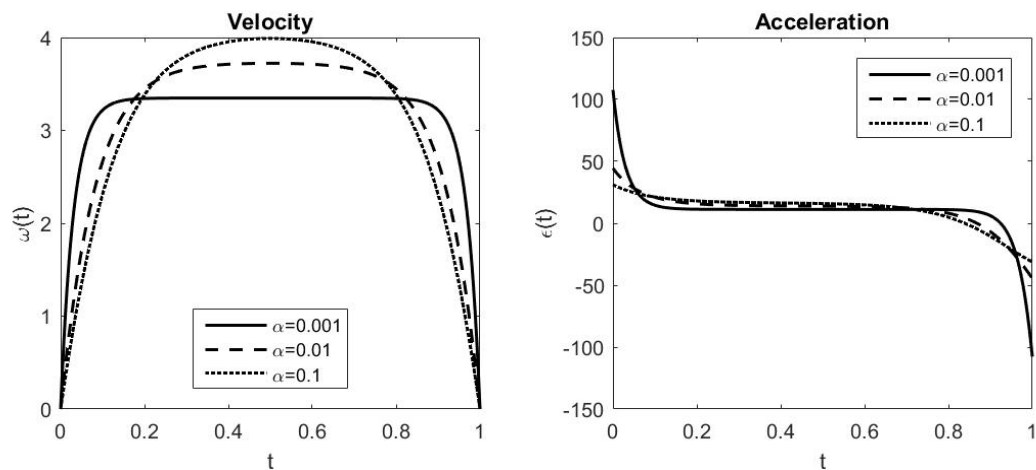

Figure 4 – Angular velocity and acceleration at different weights α for inertia forces at the optimal mode of movement along the semi-circle

Figure 3 shows the law of rotation angle change in the interval [0, π] for movement along the semicircle for different values of weight α in the cost functional. Increasing α leads to a smoother mode of movement. Figures 4 show the corresponding profiles of angular velocities and accelerations for the optimal solution.

Example 3. The trajectory is described by a parabola, for example,

$$x = p, y = kp^2 + b,$$
$$x' = 1, y' = 2kp,$$
$$x'' = 0, y'' = 2k.$$

In this case, equation (29) together with (30) is reduced to a canonical form:

$$\begin{cases} \dot{z}_0(t) = z_1(t), \\ \dot{z}_1(t) = z_2(t), \\ \dot{z}_2(t) = z_3(t) \\ \dot{z}_3 = \dfrac{1}{\alpha m} z_2 - \dfrac{4k^2 z_0}{\alpha m(1+4k^2)}(z_1 - 2z_1^2) - \dfrac{4k^2 z_0}{(1+4k^2)}(4z_1 z_3 + 3z_2^2). \end{cases}$$

Example 4. The trajectory is an ellipse:

$$\begin{aligned} & x = a\cos p, \; y = b\sin p, \\ & x' = -a\sin p, \; y' = b\cos p, \\ & x'' = -a\cos p, \; y'' = -b\sin p, \\ & x''' = a\sin p, \; y''' = -b\cos p, \\ & x^{(IV)} = a\cos p, \; y^{(IV)} = b\sin p. \end{aligned}$$

Given the notation (30), we get a system of quasilinear equations:

$$\begin{cases} \dot{z}_0(t) = z_1(t), \\ \dot{z}_1(t) = z_2(t), \\ \dot{z}_2(t) = z_3(t), \\ \dot{z}_3 = \dfrac{1}{\alpha m} z_2 - \dfrac{(a^2-b^2)\sin p \cos p}{(a^2 \sin^2 p + b^2 \cos^2 p)} \left( \dfrac{z_1 - 2z_1^2}{\alpha m} + 4z_1 z_3 + 3z_2^2 \right) \\ \quad + 6z_1^2 z_2 + \dfrac{5(a^2-b^2)\sin p \cos p}{(a^2 \sin^2 p + b^2 \cos^2 p)} z_1^4. \end{cases} \quad (31)$$

For the numerical solution of the system (31), it should be completed with four boundary conditions. Here velocities and positions at the initial and final time moments are given. This means that the boundary values of the functions are specified:

$$\begin{aligned} & z_0(0) = p_0, \\ & z_1(0) = 0, \\ & z_0(1) = p_1, \\ & z_1(1) = 0. \end{aligned} \quad (32)$$

As a solution to the minimization problem, we get the function $p(t) = z_0(t)$, which defines the parameter change along the trajectory over time. As a result, we get the law of coordinate change $x = x(p(t))$, $y = y(p(t))$. Recall that in all numerical examples, the movement interval T was used as a time scale.

Let in (31)-(32) the movement be made along an elliptical trajectory with parameters a = 1, b = 2. Let's take the polar angle in the plane (x, y) as the trajectory parameter.

Figure 5 on the left shows the tool path. The small circle indicates the initial position, which is also the final one. The right figure 5 shows the optimal solutions of p(t) for different values of the weight α, which is responsible for the inertia forces. For the most comfortable movement of the tool, this parameter should be greater.

Figures 6 and 7 show the velocity and acceleration profiles, respectively, corresponding to the optimal mode of movement along this trajectory. It is seen that the optimal velocity mode differs significantly from the traditional trapezoidal velocity profile.

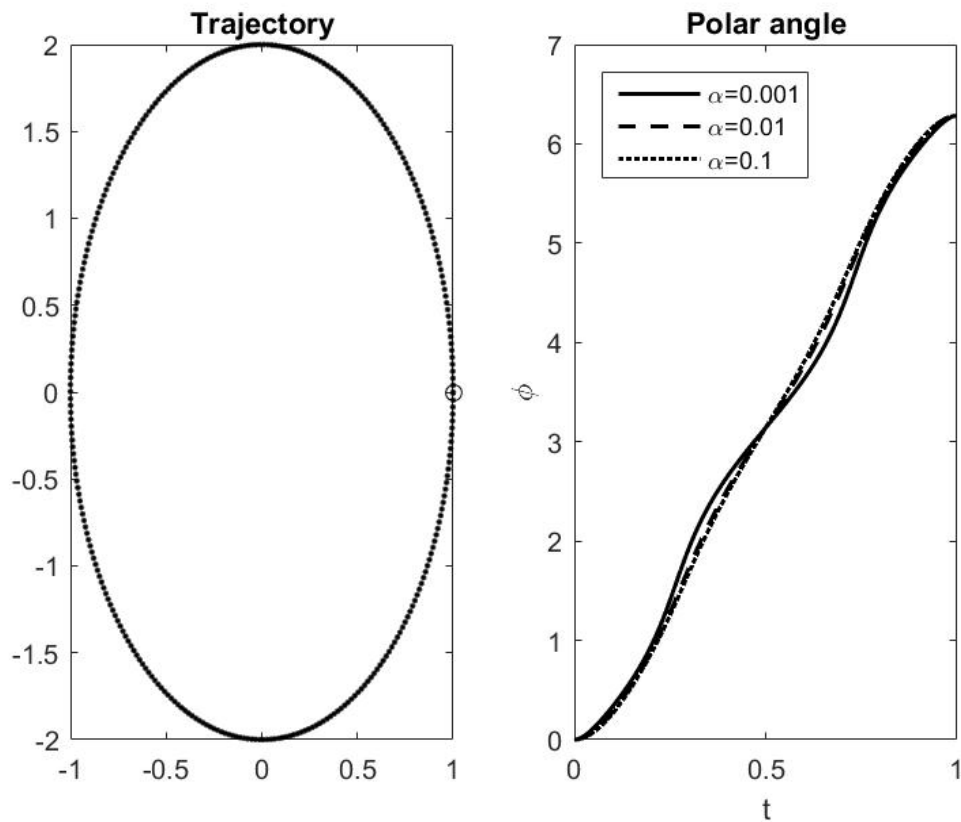

Figure 5 – an example of a specified toolpath is an ellipse (on the left);

the optimal law of polar angle change in the interval $[0.2\pi]$ for different regularization parameters (on the right).

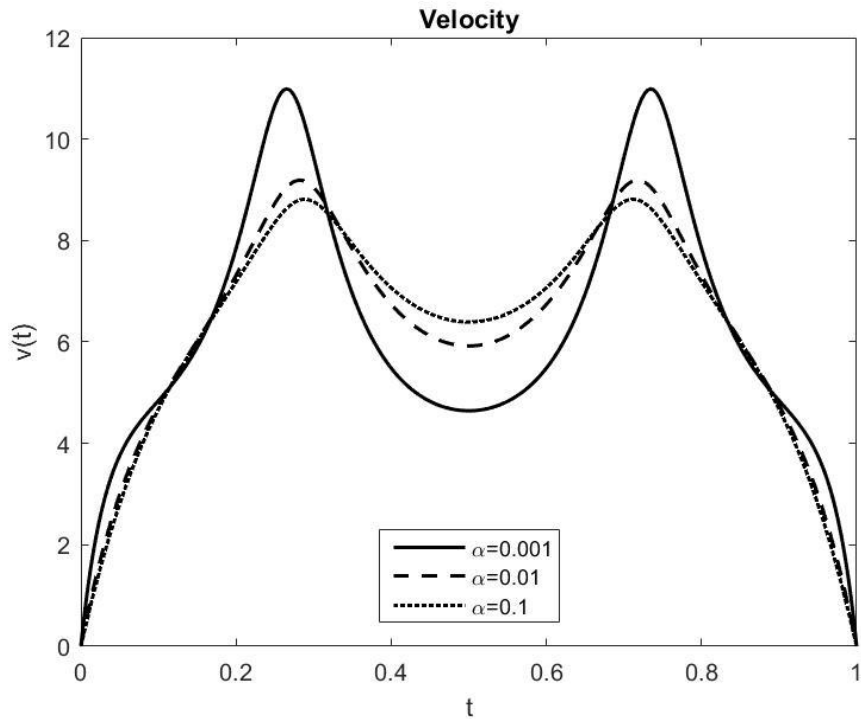

Figure 6 – The law of angular velocity change for optimal mode of movement along the trajectory for different weights of inertia forces in the quality functional

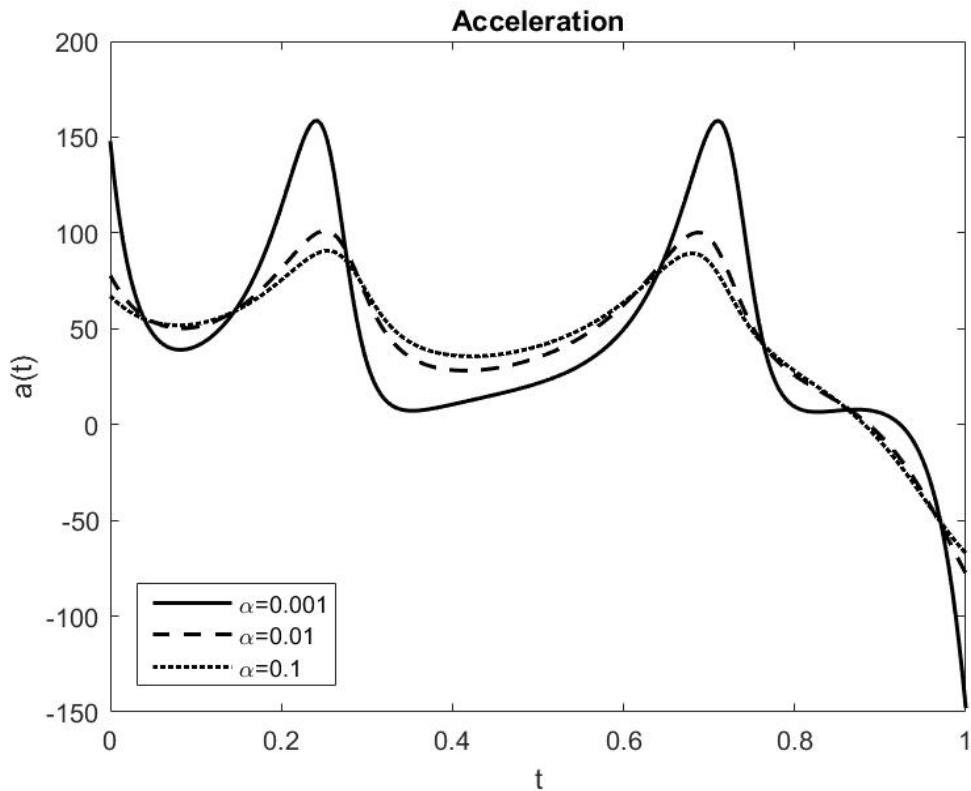

Figure 7 - The law of angular acceleration change for different weight values α in the quality functional

As it can be seen from Figure 7, by increasing the parameter α, it is possible to reduce the differences in acceleration of tool movement, which improves the comfort of movement and the influence of inertia forces

**Conclusion**

For robots that performs the same repetitive movements, moving along a given trajectory, the law of motion of the working tool can be calculated once in advance and then implemented according to a given table of successive tool positions.  Therefore, despite the complex, nonlinear nature of the resulting equations, solutions to these equations can be built in advance, regardless of the technical implementation of the robot, based on available numerical methods.

We have shown what the preferred law of motion should be for the examples of straight, circular, parabolic, and elliptical trajectories over a given time interval to minimize kinetic energy and weighted inertia. In the case of an arbitrary trajectory defined by a table of points, it is possible to generalize the proposed method by uniformly approximating the trajectory with analytical expressions. However, in this case, it will be necessary to solve system (29) - (30), which has a rather complicated form.

Note that since we are limited to considering only the movement of the robot's center of gravity, the approach described above is universal for all robots that implement movements along a planar trajectory. In particular, the same consideration can be applied to the DexTar robot [1] - [2]. The presented optimization method of the movement law can also be used in 3D printing tasks. For example, when the printed layer is completely filled with polymer material, the printer makes many movements along straight lines parallel each other. Optimization of the straight line movement will reduce the forces of inertia. It will naturally also be necessary to control the supply of material, which is regulated by the current speed of the device head movement.  But this problem is still beyond our consideration.

The subsequent problem of implementing the obtained law of motion depends on the robot's device and requires knowledge of the equations of motion for each individual robot.

The work is supported by Ministry of Education and Science of Republic of Kazakhstan, the agreement number AP05133190.

1. Patle, B.K., Babu L, G., Pandey, A., et al. A review: On path planning strategies for navigation of mobile robot.
2. Rosyid, A., El-Khasawneh, B. Alazzam, A. Review article: Performance measures of parallel kinematics manipulators.
3. Moradi, M.; Naraghi, M.; Nikoobin, A. Indirect Optimal Trajectory Planning of Robotic Manipulators with the Homotopy Continuation Technique. 2nd RSI/ISM International Conference on Robotics and Mechatronics (ICRoM): Tehran, IRAN: Oct 15-17, 2014
4. C.M.Gosselin,J.Angeles,A globe performance index for the kinematic optimization of robotic manipulators,ASME Journal ofMechanicalDesign113(3) (1991)220–226.
5. T.Huang, Z.X.Li, M.Li, D.G.Chetwynd, C. M. Gosselin, Conceptual design and dimensional synthesis of a novel2-DOF translational parallel robot for pick-and-place operations, ASME Journal ofMechanicalDesign126(3)(2004) 449–455.
6. T.Huang, M.Li, Z.X.Li, D.G.Chetwynd,D. J. Whitehouse, Optimal kinematic design of 2-DOF parallel manipulators with well-shaped workspace bounded by a specified conditioning index, IEEE Trans.on Robotics andAutomation20(3) (2004)538–543.
7. K.Miller,Optimal design and modeling of spatial parallel manipulators, International Journal of Robotics Research 23(2)(2004) 127–140.
8. M.Stock, K.Miller,Optimal kinematic design of spatial parallel manipulators: application to linear delta robot,ASME Journal ofMechanicalDesign125(2) (2004)292–301.
9. V.Nabat, M.O.Rodriguez, O. Company, S. Krut,F.Pierrot,Par4: very high speed parallel robot for pick-and-place,Proceedings of the IEEE/RSJ International Conference on Intelligent Robots and Systems,Alberta, Canada, 2005,1202–1207.

10. X.J.Liu, J. S.Wang, A new methodology for optimal kinematic design of parallel mechanisms,Mechanismand Machine Theory42(9) (2007) 1210–1224.
11. X.J.Liu, Q.M. Wang,J. S.Wang, Kinematics, dynamics and dimensional synthesisof a novel 2-DOFtranslational manipulator,Journal of Intelligent and Robotic Systems41(2004) 205–224.[12]M.A.Laribia,L.Romdhanea,S.Zeghloulb
12. Gharaaty, Sepehr; Shu, Tingting; Joubair, Ahmed, et al. Online pose correction of an industrial robot using an optical coordinate measure machine system: International journal of advanced robotic systems. Vol. 15(4), ID: 1729881418787915. - 2018
13. Hao Qi; Guan Liwen; Wang Jianxin. Dynamic Performance Evaluation of a 2-DoF Planar Parallel Mechanism, International journal of advanced robotic systems, Vol. 9,- ID: 250. - 2012 .
14. Huang, Tian; Liu, Songtao; Mei, Jiangping et al. Optimal design of a 2-DOF pick-and-place parallel robot using dynamic performance indices and angular constraints. Mechanism and machine theory, Vol.70, - p. 246-253, - 2013
15. Kim, Han Sung, Dynamics Analysis of a 2-DOF Planar Translational Parallel Manipulator. Journal of Manufacturing Engineening & Technology, - Vol. 22(2), p.185-191. - 2013
16. Hao Qi; Guan Liwen; Wang Jianxin, Dynamic Performance Evaluation of a 2-DoF Planar Parallel Mechanism. Inernational journal of advanced robotic systems, Vol. 9 ID 250,  - 2012
17. Sergiu-Dan Stan, Vistrian Mătieş and Radu Bălan Optimization of a 2 DOF Micro Parallel Robot Using Genetic Algorithms. -Technical University of Cluj-Napoca Romania
18. D.I. Malyshev, M.A.Posypkin, L.A.Rybak, A.L.Usov. - Analysis of the robot 's work area DexTAR dexterous twin-arm robot. - International Journal of Open Information Technologies ISSN: 2307-8162 vol. 6, no.7, 2018.
19. Merlet J. P. Parallel robots. – Springer Science & Business Media, 2006. – T.128.
20. K Kolmogorov A.N., Fomin S.V. Elements of the Theory of Functions and Functional Analysis. - 7th ed. - M .: Fizmatlit, 2004. — P.572. — ISBN 5-9221-0266-4.
21. Sobolev S.L. Some applications of functional analysis in mathematical physics. - Moscow: Nauka, 1988. ─ P.333.
22. Vasilyev F.P. Methods of solving extreme problems. - M.: Nauka, 1981. – P.400.